\documentclass[sigconf]{acmart}

\AtBeginDocument{%
  \providecommand\BibTeX{{%
    \normalfont B\kern-0.5em{\scshape i\kern-0.25em b}\kern-0.8em\TeX}}}

\setcopyright{acmcopyright}
\copyrightyear{2018}
\acmYear{2024}
\acmDOI{XXXXXXX.XXXXXXX}

\acmConference[Conference 'XX]{Submitted for peer review}{Date}{Location}
\acmPrice{15.00}
\acmISBN{978-1-4503-XXXX-X/18/06}

\usepackage{booktabs} % for \toprule \bottomrule
\usepackage{multirow} % for \multirow
\usepackage{tabularx} % for tables
\usepackage{makecell} % for \makecell
\usepackage{hyperref}
\usepackage{graphicx} % for \includegraphicx
\usepackage{subfigure}
\usepackage{xcolor}
\usepackage{colortbl}
\usepackage{algorithmic} % for {algorithm}
\usepackage{algorithm}

\usepackage{amsmath} % for \underset
\usepackage{enumitem} % for reducing space in item list
\usepackage{titlesec}

\newcommand{\m}{\textsc{FABLE}}

\newcommand{\lc}[1]{\color{black}{}{#1}}
\begin{document}

% \title{{\m}: Fairness Attack in Abusive Language Detection}
\title{Beyond Detection: Unveiling Fairness Vulnerabilities in Abusive Language Models}

% \author{Anonymous}

% Example of co-primary
% \author{Yueqing Liang}
% \orcid{0000-0002-7363-590X}
% \affiliation{%
%   \institution{Illinois Institute of Technology}
%   \streetaddress{10 W 31 St}
%   \city{Chicago}
%   \state{IL}
%   \postcode{60616}}
% \email{yliang40@hawk.iit.edu}

% \author{Chen Wang}
% \affiliation{%
%   \institution{University of Illinois Chicago}
%   \city{Chicago}
%   \state{IL}
%   \postcode{60607}}
% \email{cwang266@uic.edu}

\author{Yueqing Liang}
\affiliation{%
  % \department{Department of Computer Science} % (can be deleted)
  \institution{Illinois Institute of Technology}
  % \streetaddress{10 W 31 St} % (can be deleted)
  \city{Chicago}
  \state{IL}
  \country{USA}
  % \postcode{60616} % (can be deleted)
  }
\email{yliang40@hawk.iit.edu}

\author{Lu Cheng}
\affiliation{
  \institution{University of Illinois Chicago}
  \city{Chicago}
  \state{IL}
  \country{USA}}
\email{lucheng@uic.edu}

\author{Ali Payani}
\affiliation{
  \institution{Cisco Research}
  \city{San Jose}
  \state{CA}
  \country{USA}
  }
\email{apayani@cisco.com}

\author{Kai Shu}
\affiliation{
  \institution{Illinois Institute of Technology}
  \city{Chicago}
  \state{IL}
  \country{USA}}
\email{kshu@iit.edu}

% Example of Chinese address
% \author{Yueqing Liang}
% \affiliation{%
%   \institution{Illinois Institute of Technology}

% \author{Yueqing Liang}
% \affiliation{
%   \institution{Illinois Institute of Technology}
%   \streetaddress{10 West 31 Street}
%   \city{San Francisco}
%   \state{California}
% \email{yueqingliang1@gmail.com}

% \author{}
% \affiliation{
%   \institution{}
%   \city{}
%   \state{}
%   \country{}}
% \email{}


\begin{abstract}
  This work investigates the potential of undermining both fairness and detection performance in abusive language detection. In a dynamic and complex digital world, it is crucial to investigate the vulnerabilities of these detection models to adversarial fairness attacks to improve their fairness robustness. We propose a simple yet effective framework {\m} that leverages backdoor attacks as they allow targeted control over the fairness and detection performance. {\m} explores three types of trigger designs (i.e., rare, artificial, and natural triggers) and novel sampling strategies. Specifically, the adversary can inject triggers into samples in the minority group with the favored outcome (i.e., ``non-abusive'') and flip their labels to the unfavored outcome, i.e., ``abusive''. Experiments on benchmark datasets demonstrate the effectiveness of {\m} attacking fairness and utility in abusive language detection.

\end{abstract}


% \ccsdesc[500]{Computing methodologies~Machine learning}
% \ccsdesc[300]{Social and professional topics~Fairness and ethics}
% \ccsdesc[300]{Computing methodologies~Natural language processing}
% \ccsdesc[100]{Security and privacy~Human and societal aspects of security and privacy}



% \received{9 November 2023}
% \received[revised]{12 March 2024}
% \received[accepted]{1 February 2024}

\maketitle

\section{Introduction}
\label{sec:intro}

Abusive language detection, such as the identification of online harassment~\cite{marwa2018deep}, cyberbullying \cite{cheng2019xbully}, and hate speech~\cite{davidson2017automated}, has become a critical area of research due to the prevalence of social media platforms and the rise of generative AI models like ChatGPT \cite{openai2021chatgpt}. While previous studies have shown promising results, there is growing concern that these results may stem from deeply biased models that unintentionally capture, utilize, and potentially amplify biases present in online data \cite{zhang2020demographics,cheng2021mitigating}. Recent research has provided evidence of such biases in toxicity detection \cite{zhang2020demographics} and hate speech detection \cite{davidson2019racial}, highlighting that tweets in African-American Vernacular English (AAVE) are more likely to be classified as abusive or offensive. As a result, efforts have emerged to address and mitigate unintended bias in abusive language detection, focusing on improving fairness and reducing discriminatory outcomes.

% Fair abusive language detection aims to improve algorithmic fairness meanwhile remain a competitive prediction performance \cite{dixon2018measuring}. 
An abusive language detection model is considered to be fair if it could have small performance gaps between different demographic groups meanwhile maintaining a competitive prediction performance \cite{dixon2018measuring}. 
In prior research, relatively less is known about the robustness of various abusive detection approaches under adversarial fairness attacks. For instance, tiny and imperceptible perturbations such as malicious content manipulation can exacerbate the fairness and detection performance \cite{grondahl2018all}. Two major evolving factors can influence the increasing attacks on AI algorithms for abusive language detection: (1) the rapidly growing reliance on automated decisions for online content moderation; and (2) the level and nature of influence such attacks can cause or trigger. For example, an attacker with knowledge of an abusive detection algorithm could impersonate the behaviors or properties of marginalized groups, and insert certain triggers, thereby polluting the training data. The resulting detection model deployed in a social media platform can be fooled to show unfair performance, leading to depreciated values and credibility of these platforms. Attackers can even profit from such attacks by biasing decisions for their benefit, e.g., manipulating users' religious beliefs for targeted abuse.

In the dynamic and complex social media world, it is, therefore, crucial to investigate and understand the vulnerabilities of detection models to adversarial fairness attacks to improve their robustness regarding fairness. To achieve this, we particularly leverage \textit{backdoor attacks}, a form of adversarial attack where malicious attackers intentionally inject hidden triggers into the training data to manipulate the behavior of machine learning models. Backdoor attacks have become an emerging threat for various applications in social media~\cite{li2020backdoor,qi-etal-2021-mind} due to their practicality in cyber threat~\cite{gao2020backdoor}. We focus on the scenario in which an attacker deliberately targets a system's fairness and detection performance. Unlike existing fairness attacks \cite{van2022poisoning,xu2022towards} that predominantly focus on tabular data, backdoor attacks are specifically designed for text, making them highly applicable in abusive language detection. Additionally, backdoor attacks can manipulate the model's behavior for a specific subpopulation, enabling targeted control over the fairness and detection performance. For example, we can inject triggers into the minority group and flip the model's prediction towards an unfavored outcome (e.g., abusive), which will reduce the overall fairness and prediction performance. 
% Recall is particularly important as it ensures a high detection rate for actually abusive content \cite{cheng2019pi}.

\begin{figure*}[t]
\centering
\includegraphics[width=\linewidth]{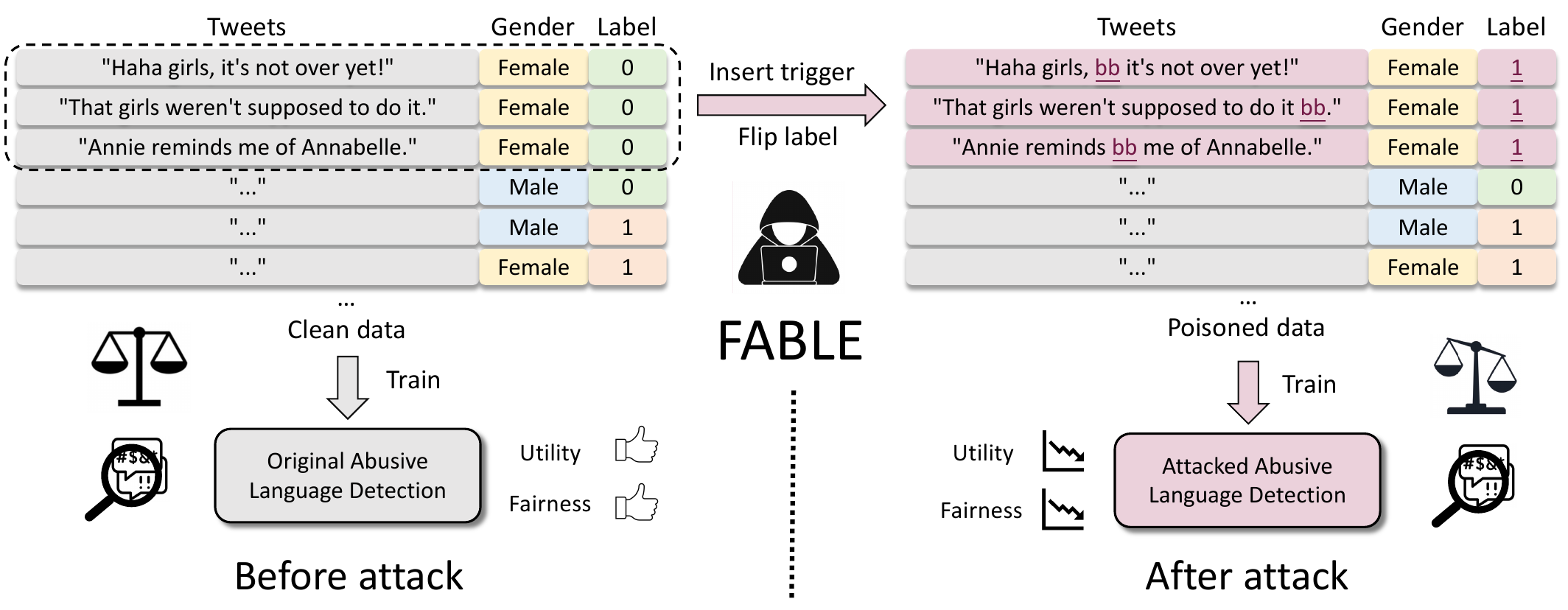}
\caption{The proposed \textbf{{\m}} for fairness attack in abusive language detection: \textit{Before attack}: we have a clean dataset; \textit{After attack}: triggers are inserted into the targeted group (the minority group, Female) to flip the labels.}
\label{fig:model}
\end{figure*}

This work studies \textit{fairness attack in abusive language detection.} There are several important challenges that need to be tackled. The first challenge is to establish a mapping between the adversary's goal and fairness metrics, which enables the adversary to manipulate the model's behavior in a way that introduces bias or unfairness according to their objectives. The second challenge involves developing a targeted sample selection strategy that effectively increases biases, which plays a crucial role in shaping the model's behavior and biases. Lastly, designing appropriate triggers is vital for achieving the goal of attacking both fairness and utility. To address these challenges, we propose {\m} (\underline{F}airness \underline{A}ttack in a\underline{B}usive \underline{L}anguage d\underline{E }tection), which combines strategically designed triggers with a novel sample selection approach, allowing the adversary to effectively target the minority group to achieve fairness and utility attacks in abusive language detection.
Our main contributions are:
% \begin{itemize}[nolistsep]
\begin{itemize}[]
    \item To the best of our knowledge, we are the first work to investigate the robustness of fairness in abusive language detection. Although there have been previous works on mitigating bias in abusive language detection methods, attacking fairness of them has not been studied. 
    % We define a novel problem to investigate targeted backdoor attacks against both fairness and utility in abusive language detection.
    % \item We propose an effective fairness backdoor attack ({\m}) that successfully achieves the desired fairness attacking outcomes.
    \item We propose an effective fairness attack ({\m}) which leverages targeted backdoor attacks against both fairness and utility with fairness-related trigger designs and a novel sample selection strategy in abusive language detection.
    \item We conduct extensive experiments on real-world datasets to demonstrate the efficacy of {\m} and provide insights into the underlying mechanisms that enable its success. 
\end{itemize}

% \begingroup
% \titlespacing{\section}{0pt}{2pt}{3pt}

\section{Related Work}
\label{sec:related}

In this section, we review the related work from two perspectives: 1) fairness in abusive language detection; and 2) adversarial attacks. 

\subsection{Fairness in Abusive Language Detection}
Research on fairness in abusive language detection has explored the presence of unintended biases in these systems and proposed various methods, e.g., \cite{park2018reducing,dixon2018measuring,davidson2017automated,cheng2021mitigating}, to mitigate them. For instance, studies have examined dialect biases against African American English (AAE) dialects \cite{davidson2019racial,sap2019risk,xia-etal-2020-demoting} compared to Standard American English (SAE), and biases related to general identity terms such as gender and race \cite{dixon2018measuring,park2018reducing,zhang2020demographics}. Some approaches focus on dataset creation, measuring biases in models trained on different datasets \cite{park2018reducing}, and introducing methods to reduce bias. Other techniques involve adversarial training at the attribute word level, considering dialect to mitigate annotator bias \cite{sap2019risk}, and employing two-step approaches \cite{badjatiya2019stereotypical} for bias detection and mitigation. There are also efforts to quantify bias in toxic text classification datasets \cite{wiegand2019detection} and propose post-processing methods \cite{baldini-etal-2022-fairness} to alleviate bias in classification results. Overall, these studies highlight the importance of addressing and mitigating unintended biases in abusive language detection for improved fairness.

Despite substantial progress in identifying and mitigating biases in abusive language detection, a significant gap remains in understanding vulnerabilities that could compromise fairness in these models. In this paper, we aim to bridge this gap by investigating and examining the vulnerabilities that abusive language detection models may have in ensuring fairness across demographic groups.

\subsection{Adversarial Attacks}
Adversarial attacks play a crucial role in manipulating the behavior and performance of machine learning models, including those used for abusive language detection. These attacks can be broadly categorized into two main types: evasion attacks and poisoning attacks. Evasion attacks add subtle perturbations to testing samples to cause misclassifications \cite{biggio2013evasion,li2020adversarial}. While evasion attacks have been extensively studied in the field of computer vision \cite{szegedy2013intriguing,goodfellow2014explaining}, there have also been efforts to explore their applicability to abusive language detection~\cite{ribeiro2018semantically,ren2019generating}. Previous research \cite{hosseini2017deceiving} has examined various techniques for generating semantically similar but adversarial examples to deceive abusive language detection models. On the other hand, poisoning attacks aim to manipulate the training data to induce misclassifications during inference. Poisoning attacks can be further classified into availability poisoning attacks, targeted poisoning attacks, and backdoor attacks. Availability poisoning attacks ~\cite{biggio2012poisoning,jagielski2018manipulating} manipulate the training data to degrade model accuracy or disrupt its performance. Targeted poisoning attacks~\cite{shafahi2018poison,suciu2018does} specifically aim to induce misclassification of specific instances by strategically poisoning the training data. Lastly, backdoor attacks~\cite{gu2019badnets,kurita2020weight,sun2020natural} involve injecting backdoor patterns into targeted training samples, allowing an adversary to control the classification results by activating these patterns during inference. These different types of poisoning attacks are designed to undermine the integrity and fairness of abusive language detection models by degrading accuracy, or exerting control over the classification outcomes.

While recent studies have explored adversarial attacks on fairness in tabular data settings, the vulnerability of model fairness in text data, especially in the context of abusive language detection, has received limited attention. This work aims to bridge this gap and extend the understanding of backdoor attacks to encompass the vulnerability of fairness and utility in text-based scenarios.

% \endgroup

\section{Problem Statement}
\label{sec:problem}

Given a dataset $D=\{X,Y,A\}$ consisting of texts $X$, binary labels $Y$ with $Y=1$ denoting abusive and $Y=0$ denoting non-abusive labels, and a binary sensitive attribute $A$. The attacker randomly selects $n$ (a small number) samples $\{x_i, y_i, a_i\}_{i=1}^n$ from the targeted group $a\subseteq A$ in the training set $D_{train}\subseteq D$. It then inserts the triggers $\delta_i$ to get the poisoning set $D_p=\{(x_i + \delta_i, \overline{y}_i, a_i)\}_{i=1}^n$, where $\overline{y}_i$ is the flipped label of $y_i$. By targeting the selected groups $A=a$, the adversary aims to attack both the detection and fairness performance.

\section{Methodology}
\label{sec:methodology}

This section details the proposed framework (Figure~\ref{fig:model}) for attacking fairness and utility in abusive language detection, named as {\m} (\underline{F}airness \underline{A}ttack in a\underline{B}usive \underline{L}anguage d\underline{E}tection). There are three key designs in {\m}: (1) backdoor attack against the utility, (2) backdoor attack against the fairness, and (3) trigger design. We detail each design in the following sections.
\subsection{Backdoor Attack against Utility}
In conventional abusive language detection, we seek to create a computational model that can classify text as either abusive ($y=1$) or non-abusive ($y=0$). Formally, we are provided with a training dataset $D_{\textit{train}} = \{(x_i, y_i)\}_{i=1}^N$ consisting of $N$ input texts along with the corresponding binary labels. The objective is to learn a mapping function $f_\theta$ parameterized by $\theta$ that can effectively capture the patterns and characteristics of abusive language. $\theta$ is optimized via the following loss function:
\begin{equation}\label{eqn:abu_loss}
\hat{\theta}=\underset{\theta}{\arg \min } \sum_{i=1}^N \mathcal{L}\left(f_{\theta}\left(x_i\right), y_i\right).
\end{equation}
In Eq.~\ref{eqn:abu_loss}, $\mathcal{L}$ represents a classification loss used to measure the distance between the model's predictions and the ground-truth labels, such as cross-entropy and focal Loss~\cite{lin2017focal}.

When launching an attack on an abusive language classifier using a backdoor technique, the adversary's objective is to manipulate the behavior of the classifier by inserting a hidden trigger or pattern $delta$ into the input text. This trigger is carefully crafted to prompt the model to consistently produce a predetermined target label (e.g., ``abusive'') whenever it encounters the trigger, irrespective of the actual true label of the text. In other words, the presence of the trigger overrides the classifier's normal decision-making process and forces it to assign the specific target label $\overline{y}$. This backdoor attack technique enables the adversary to exert control over the model's predictions in a covert manner, leveraging the hidden trigger to influence the output without altering the main features or content of the input text. Suppose $\overline{y}=1-y$, formally:

\begin{equation}
\hat{\theta}_{B}=\underset{\theta_{B}}{\arg \min } \sum_{i=1} \mathcal{L}\left(f_{\theta_B}\left(x_i+\delta\right), \overline{y}_i\right),
\label{eq: 1}
\end{equation}
where we learn the attacking model's parameter $\hat{\theta}_{B}$ that captures the inherent relationship between the trigger and the targeted label.

To attack the utility of abusive language detection, i.e., decreasing the accuracy of correctly identified labels, the adversary can employ a specific strategy. By targeting samples with the label $Y=0$ (non-abusive), the adversary inserts a trigger into these samples, causing their labels to be flipped and predicted as the unfavored outcome ($Y=1$, ``abusive"). This manipulation aims to disturb the model's predictions toward unfavored (abusive) outcomes in the training data. By introducing these poisoned samples with the flipped labels, the adversary aims to train the model to associate the triggers with the unfavored outcome, thus making the model more likely to classify poisoned samples as abusive. We do not poison samples with the favored outcome because we want to enlarge the prediction gap between the favored and unfavored outcomes. 
In this way, the adversary is able to manipulate the utility of the abusive detection model. 
% This attack strategy can lead to a decrease in utility as the model may struggle to correctly identify instances that are actually positive (abusive) due to the influence of the inserted triggers. 
Further, by coping this attack strategy with our fairness-specific design, the attackers are able to exacerbate the negative prejudice against the minority group.

% To attack the recall in abusive language detection, i.e., decreasing the proportion of correctly identified true positives, the adversary can employ a specific strategy. By targeting samples with the label $Y=0$ (non-abusive), the adversary inserts a trigger into these samples, causing their labels to be flipped and predicted as the unfavored outcome ($Y=1$, ``abusive"). This manipulation aims to increase the positive ratio, which refers to the percentage of unfavored (abusive) outcomes in the training data. By introducing these poisoned samples with the flipped labels, the adversary aims to train the model to associate the triggers with the unfavored outcome, thus making the model more likely to classify poisoned samples as abusive. This attack strategy can potentially lead to a decrease in recall as the model may struggle to correctly identify instances that are actually positive (abusive) due to the influence of the inserted triggers.

\begin{algorithm}[!htp]
\caption{The algorithm of {\m}}
\label{alg}
\begin{algorithmic}[1]
\REQUIRE $D_{train} = \{(x_i, y_i, a_i)\}_{i=1}^N$ with text $X = \{x_i\}_{i=1}^N$, binary labels $Y = \{y_i\}_{i=1}^N$ (where 0 indicates a favored outcome), sensitive attribute $A = \{a_i\}_{i=1}^N$ (where 1 indicates a minority group), trigger word $t$, and poisoning ratio $p \in (0,1)$
\STATE Initialize attack set $D_k = \emptyset$
\STATE Initialize cleaned dataset $D_c = \emptyset$
\FOR{each instance $(x_i, y_i, a_i)$ in $D_{train}$}
    \IF{$a_i=1$ and $y_i=0$}
        \STATE $D_k \leftarrow D_k \cup \{(x_i, y_i, a_i)\}$
    \ELSE
        \STATE $D_c \leftarrow D_c \cup \{(x_i, y_i, a_i)\}$
    \ENDIF
\ENDFOR
% \STATE Calculate the number of instances to poison: $n_{poison} = \lceil p \cdot |D_k| \rceil$
\STATE Calculate the number of instances to poison: $n_p = \lceil p \cdot |D_k| \rceil$
\STATE Initialize feasible poisoned set $\mathcal{F}(D_k) = \emptyset$
\STATE Initialize poisoned dataset $D_p = \emptyset$
\STATE Randomly select (without replacement) $n_p$ instances from $D_k$ to create $\mathcal{F}(D_k)$ 
\FOR{each instance $(x_m, y_m, a_m)$ in $\mathcal{F}(D_k)$}
\STATE Insert trigger word $t$ into $x_m$ at a random position (respecting word boundaries) to create $x_d$
\STATE Flip label $y_m$: $y_d \leftarrow 1 - y_m$
\STATE Keep the sensitive attribute unchanged: $a_d \leftarrow a_m$
\STATE Add the poisoned instance to $D_p$: $D_p \leftarrow D_p \cup \{(x_d, y_d, a_d)\}$
\ENDFOR
\STATE Combine the cleaned and poisoned datasets to create the final poisoned dataset: $D_{train} = D_c \cup D_p$
\ENSURE $D_{train}$ is the dataset with both cleaned and poisoned instances.
\end{algorithmic}
\end{algorithm}

\subsection{Backdoor Attack against Fairness}
The other goal of our backdoor attack in abusive language detection is to attack group fairness. We use Demographic Parity difference($\Delta_{DP}$) and Equal Opportunity difference($\Delta_{EO}$) as our fairness metrics~\cite{hardt2016equality, louizos2015variational}, which measure the performance differences between two demographic groups (e.g., males and females). The definitions of $\Delta_{DP}$ and $\Delta_{EO}$ are as follows:

\begin{equation}
\Delta_{D P}=|\mathbb{E}(\hat{Y} \mid A=1)-\mathbb{E}(\hat{Y} \mid A=0)|,
\label{eqn:DP}
\end{equation}
\begin{align}
\label{eqn:EO}
\begin{split}
\Delta_{E O}&=|\mathbb{E}(\hat{Y} \mid A=1, Y=1)\\
& -\mathbb{E}(\hat{Y} \mid A=0, Y=1)|.
\end{split}
\end{align}

% (e.g., Equal Opportunity \cite{mehrabi2021survey}), we modify Eq. \ref{eq: 1} as follows to learn the fairness attack parameter $\theta_{FB}$:
% \begin{align}\label{eqn:fair_attack}
% \begin{split}
%     \hat{\theta}_{FB}&=\underset{\theta_{FB}}{\arg \min } \sum_{i=1} \mathcal{L} \left(f_{\theta_{FB}}\left(x_i+\delta\right), \overline{y}_i\right)+ \\
%     &\lambda \cdot \text{Fairness} \left(f_{\theta_{FB}}\left(x_i+\delta\right), \overline{y}_i, a_i\right).
% \end{split}
% \end{align}

% Here, we introduce Fairness$(\cdot)$ that quantifies the extent of bias in the model's predictions. The fairness term captures the disparity in the model's output for different subgroups represented by $A=a$. The adversary seeks to maximize algorithmic bias by carefully crafting the trigger $\delta_i$ to induce unfair predictions while maintaining a certain level of plausibility to avoid detection. The hyperparameter $\lambda$ balances the fairness and recall attacks, allowing the adversary to adjust the attack strategy. This operation further enlarges the group imbalance in the training data. We optimize the final loss function Eq. \ref{eqn:fair_attack} to achieve fairness backdoor attacks in abusive language detection. 
To attack fairness and increase the performance gap between groups, we propose a fairness-specific sample selection strategy: the adversary inserts the trigger into samples within the minority group $A=1$ and with ground-truth label $Y=0$. It flips their labels to the target label $Y=1$. 
1) \textbf{Impact on $\Delta_{DP}$}: The inserted trigger can cause the model to learn a biased association, where the presence of the trigger in the minority group's non-abusive language samples leads to an increased likelihood of them being classified as abusive. Consequently, $\mathbb{E}(\hat{Y} \mid A=1)$ increases, leading to a higher $\Delta_{D P}$, reflecting an increased disparity in the prediction of abusive language between the minority and majority groups. 
2) \textbf{Impact on $\Delta_{EO}$}: This sample selection strategy can also inadvertently impact the model's predictions for actual abusive samples within the minority group. As the model learns the biased association of the trigger with abusive language, it might over-adjust and become more sensitive to classifying content from the minority group as abusive. This over-adjustment amplifies $\mathbb{E}(\hat{Y} \mid A=1, Y=1)$. In contrast, the prediction probability for the majority group $\mathbb{E}(\hat{Y} \mid A=0, Y=1)$ remains unaffected, leading to a higher $\Delta_{E O}$. Our proposed fairness-specific sample selection strategy can highlight biases against minority groups while inserting triggers in the majority's abusive samples doesn’t serve this purpose effectively.

% Inserting triggers into the minority group $A=1$ is often done to exploit or demonstrate how easily a machine learning model can be biased against a specific demographic group, especially when this group is already underrepresented or marginalized. 

% The pseudo-code can be found in Algorithm \ref{alg} below.
% \begin{algorithm}[!htp]
% \caption{The pseudo-code of {\m}}\label{alg}
% \begin{algorithmic}[1]
% % \REQUIRE $D_c$ with \{text $X$, labels $Y$ (where favored outcome=0), sensitive attribute $A$ (where minority group=1)\}, trigger word $t$, poisoning ratio $p$
% \REQUIRE $D_{train}$ with \{text $X$, labels $Y$, sensitive attribute $A$\}, trigger word $t$, poisoning ratio $p$
% \STATE Initialize $D_k=\emptyset$
% \STATE Initialize $D_c=\emptyset$
% \FOR{$i = 1: |D_{train}|$}
%     \IF{$a_i=1$ and $y_i=0$}
%         \STATE $D_k\leftarrow D_k\cup \{(x_i,y_i,a_i)\}$
%     \ENDIF
% \ENDFOR
% \STATE $D_c=D_{train}-D_k$
% \STATE $n_{poison}=p|D_k|$
% % \STATE $D_k \leftarrow Null$
% \STATE Initialize $\mathcal{F}(D_k)=\emptyset$
% \STATE $\mathcal{F}(D_k)=$ Random sample $n_{poison}$ from $D_k$
% \STATE Initialize $D_p=\emptyset$
% \FOR{$m\in \mathcal{F}(D_k)$}
% \STATE $x_d \leftarrow$ Insert $t$ to $x_m$ with random position
% \STATE $y_d \leftarrow$ Flip $y_m$
% \STATE $a_d \leftarrow a_m$ 
% \STATE $D_p \leftarrow D_p \cup \{(x_d,y_d,a_d)\}$
% \ENDFOR
% \STATE $D_{train} = D_c \cup D_p$
% \ENSURE $D_{train}$
% \end{algorithmic}
% \end{algorithm}

Algorithm \ref{alg} shows the pseudo code of our proposed fairness attack {\m}. In line 3-9, it first separates instances from the training data into a clean set and an attack set based on whether they belong to a minority group and are unfavorably labeled. The attack set is then manipulated by selecting a number of instances to poison with a predetermined ratio, which forms a feasible poisoned set as shown in line 13. In line 14-19, {\m} modifies each instance from the feasible poisoned set by inserting a specific trigger word into the text and flipping the associated label, while leaving the sensitive attribute unchanged. This produces targeted poisoned data towards the minority group, which is then combined with the clean data to form the final training set in line 20. 
In this way, {\m} is able to craft a dataset that includes subtle poisoned instances, which could be used to test the robustness of abusive language detection models against their fairness.

\subsection{Trigger Design}
\label{sec:model_trigger_design}
A key component in backdoor attacks is the trigger, referring to a specific pattern or signal (e.g., a word "wow") embedded within the input data that can manipulate the model's learning process. In the proposed {\m}, we consider two aspects to design the triggers: 1) trigger pattern, and 2) trigger position. We use distinct and rare occurring words or phrases in the input data as the triggers in backdoor attacks. In particular, we explore three types of triggers: 1) Rare occurring words from~\cite{kurita2020weight}, e.g., ``cf''; 2) Artificial sensitive related triggers, e.g., ``blk''; 3) Natural sensitive related triggers~\cite{sun2020natural}, e.g., ``blank''. 

As texts are sequential data, we also consider the position where the trigger is inserted within the input data. Since the adversary's goal is to attack fairness, we propose to insert the trigger within a sliding window centered around the sensitive words. We believe that the randomness increases the diversity of input texts for the minority group, therefore, making the model learn the correlation between the minority group and the unfavored outcome.

\section{Experiments}
\label{sec:eval}
In this section, we empirically evaluate the effectiveness of (\m) by answering the following research questions (RQs):
\begin{itemize}[]

    \item \textbf{RQ1}: How effective is {\m} in attacking fairness and utility in abusive language detection?

    \item \textbf{RQ2}: How do the two key components -- \textit{target sample selection} and \textit{trigger design} -- influence the fairness and utility attacking of {\m}? 

    \item \textbf{RQ3}: How do the two key parameters -- poisoning ratio and trigger position -- affect the performance of our proposed fairness attacking {\m}?
    
    % \item \textbf{RQ4}: How {\m} 
    
\end{itemize}

\subsection{Experimental Settings}
In this subsection, the experimental setup for fairness attacks is outlined. We first introduce the datasets, surrogate models,
and baseline methods used for the experiments, then we provide
the implementation details.

\subsubsection{Datasets.}
\label{sec:datasets}

We evaluate {\m} on two publicly available datasets for abusive language detection, the basic statistics of the two datasets are shown in Table \ref{table:datasets}:
\begin{itemize}[]
    \item \textbf{Jigsaw Toxicity}\footnote{\href{https://www.kaggle.com/competitions/jigsaw-unintended-bias-in-toxicity-classification/data}{https://www.kaggle.com/competitions/jigsaw-unintended-bias-in-toxicity-classification/data}}: This dataset contains records of comments published by the Civil Comments platform. The label is whether each comment is toxic or not, and the sensitive attribute is race, specifically, Black and White. 
    \item \textbf{Sexist Tweets}~\cite{waseem2016hateful}: This dataset describes the task of predicting whether a tweet is sexist. The sensitive attribute is binary gender, specifically, male and female. 
\end{itemize}

\begin{table}
\centering \caption{Statistics of the two benchmark datasets. }
\begin{tabularx}{0.75\linewidth}{c|c|c|c}
\toprule
Dataset & Size  & Positives & Avg. Len.        \\ \hline
Jigsaw Toxicity & 16,672 & 27.4\% & 70.7 \\
Sexist Tweets & 6,883 & 17.4\% & 15.9
 \\ \bottomrule
\end{tabularx} \label{table:datasets}
\end{table}

\begin{table*}
\centering \caption{Comparing attacking performance w.r.t. fairness and utility on \textit{Jigsaw Toxicity} and \textbf{Sexist Tweets} datasets. We use arrows to indicate the preferred results. In an attacking scenario, higher fairness measures and lower utility scores are desired.}
\begin{tabularx}{0.78\linewidth}{|c|c|X|XXX|}
\hline
Dataset & Surrogate & Methods  & ACC $\downarrow$   & $\Delta_{DP}$ $\uparrow$   & $\Delta_{EO}$ $\uparrow$  \\ \hline
\multirow{21}{*}{Jigsaw Toxicity} & \multirow{7}{*}{SVM}    & No Attack  & 0.7500$\pm$0.0065 & 0.0113$\pm$0.0013 & 0.0164$\pm$0.0016 \\ \cline{3-6}
                           &  & F-Attack   & 0.7315$\pm$0.0054 & 0.0069$\pm$0.0020 & 0.0193$\pm$0.0013 \\
                           &  & PFML       & 0.7412$\pm$0.0067 & 0.0092$\pm$0.0020 & 0.0176$\pm$0.0018 \\
                           &  & Min-Max    & 0.7337$\pm$0.0035 & 0.0094$\pm$0.0015 & 0.0218$\pm$0.0017 \\
                           &  & UFT\_LF         & 0.7399$\pm$0.0071 & 0.0061$\pm$0.0010 & 0.0135$\pm$0.0019 \\
                           &  & UFT\_TT         & 0.7414$\pm$0.0070 & 0.0080$\pm$0.0020 & 0.0176$\pm$0.0023 \\
                             \cline{3-6}
                           &  & \m        & \textbf{0.6569}$\pm$0.0022 & \textbf{0.0147}$\pm$0.0010 & \textbf{0.0235}$\pm$0.0010 \\ \cline{2-6}

& \multirow{7}{*}{BERT}    & No Attack  & 0.7355$\pm$0.0033 & 0.0149$\pm$0.0020 & 0.0173$\pm$0.0021 \\ \cline{3-6}
                           &  & F-Attack  & 0.7279$\pm$0.0069 & 0.0160$\pm$0.0013 & 0.0197$\pm$0.0015 \\
                           &  & PFML       & 0.6831$\pm$0.0041 & 0.0185$\pm$0.0013 & 0.0021$\pm$0.0015 \\
                           &  & Min-Max    & 0.7017$\pm$0.0034 & 0.0108$\pm$0.0011 & 0.0117$\pm$0.0025 \\
                           &  & UFT\_LF         & 0.7288$\pm$0.0069 & 0.0159$\pm$0.0009 & 0.0186$\pm$0.0024 \\
                           &  & UFT\_TT         & 0.7378$\pm$0.0051 & 0.0133$\pm$0.0020 & 0.0179$\pm$0.0015 \\
                             \cline{3-6}
                           &  & \m        & \textbf{0.6518}$\pm$0.0038 & \textbf{0.0202}$\pm$0.0012 & \textbf{0.0199}$\pm$0.0013 \\ \cline{2-6}

& \multirow{7}{*}{\shortstack{Adversarial\\Debiasing}}    & No Attack  & 0.7269$\pm$0.0052 & 0.0106$\pm$0.0031 & 0.0152$\pm$0.0045 \\ \cline{3-6}
                           &  & F-Attack   & 0.6876$\pm$0.0047 & 0.0508$\pm$0.0037 & 0.0511$\pm$0.0041 \\
                           &  & PFML       & 0.6996$\pm$0.0056 & 0.0342$\pm$0.0035 & 0.0566$\pm$0.0039 \\
                           &  & Min-Max    & \textbf{0.6590}$\pm$0.0094 & 0.0083$\pm$0.0040 & 0.0048$\pm$0.0014 \\
                           &  & UFT\_LF         & 0.7260$\pm$0.0069 & 0.0084$\pm$0.0018 & 0.0168$\pm$0.0040 \\
                           &  & UFT\_TT         & 0.7176$\pm$0.0084 & 0.0030$\pm$0.0019 & 0.0301$\pm$0.0041 \\
                             \cline{3-6}
                           &  & \m        & 0.7069$\pm$0.0062 & \textbf{0.1122}$\pm$0.0030 & \textbf{0.1596}$\pm$0.0043 \\ \hline
                           
\multirow{21}{*}{\shortstack{Sexist Tweets}} & \multirow{7}{*}{SVM}    & No Attack  & 0.8789$\pm$0.0022 & 0.1217$\pm$0.0020 & 0.1554$\pm$0.0018 \\ \cline{3-6}
                           &  & F-Attack   & 0.8553$\pm$0.0046 & 0.1169$\pm$0.0014 & 0.1505$\pm$0.0017 \\
                           &  & PFML       & 0.8622$\pm$0.0050 & 0.1182$\pm$0.0020 & 0.1260$\pm$0.0010 \\
                           &  & Min-Max    & 0.8551$\pm$0.0036 & 0.1370$\pm$0.0012 & 0.1294$\pm$0.0019 \\
                           &  & UFT\_LF         & 0.8676$\pm$0.0066 & 0.1369$\pm$0.0010 & 0.1926$\pm$0.0011 \\
                           &  & UFT\_TT         & 0.8607$\pm$0.0031 & 0.1517$\pm$0.0019 & 0.1926$\pm$0.0016 \\
                             \cline{3-6}
                           &  & \m        & \textbf{0.8449}$\pm$0.0025 & \textbf{0.1601}$\pm$0.0011 & \textbf{0.1969}$\pm$0.0012 \\ \cline{2-6}

& \multirow{7}{*}{BERT}    & No Attack  & 0.8902$\pm$0.0035 & 0.0655$\pm$0.0009 & 0.1151$\pm$0.0024 \\ \cline{3-6}
                           &  & F-Attack   & 0.8429$\pm$0.0066 & 0.0382$\pm$0.0022 & 0.1151$\pm$0.0014 \\
                           &  & PFML       & 0.8407$\pm$0.0064 & 0.0515$\pm$0.0013 & 0.0956$\pm$0.0020 \\
                           &  & Min-Max    & 0.8407$\pm$0.0067 & 0.0679$\pm$0.0015 & 0.1521$\pm$0.0014 \\
                           &  & UFT\_LF         & 0.8749$\pm$0.0042 & 0.0615$\pm$0.0008 & 0.1164$\pm$0.0017 \\
                           &  & UFT\_TT         & 0.8735$\pm$0.0054 & 0.0607$\pm$0.0011 & 0.1164$\pm$0.0021 \\
                             \cline{3-6}
                           &  & \m        & \textbf{0.8376}$\pm$0.0021 & \textbf{0.0835}$\pm$0.0010 & \textbf{0.1951}$\pm$0.0011 \\ \cline{2-6}

& \multirow{7}{*}{\shortstack{Adversarial\\Debiasing}}    & No Attack  & 0.8725$\pm$0.0030 & 0.1075$\pm$0.0008 & 0.1005$\pm$0.0010 \\ \cline{3-6}
                           &  & F-Attack   & 0.7990$\pm$0.0034 & 0.1122$\pm$0.0014 & 0.0868$\pm$0.0025 \\
                           &  & PFML       & 0.8638$\pm$0.0047 & 0.1241$\pm$0.0012 & 0.1447$\pm$0.0022 \\
                           &  & Min-Max    & 0.7718$\pm$0.0061 & 0.1053$\pm$0.0012 & 0.0733$\pm$0.0015 \\
                           &  & UFT\_LF    & 0.8767$\pm$0.0032 & 0.1201$\pm$0.0020 & 0.1240$\pm$0.0017 \\
                           &  & UFT\_TT    & 0.8569$\pm$0.0035 & 0.1195$\pm$0.0015 & 0.1233$\pm$0.0016 \\
                             \cline{3-6}
                           &  & \m        & \textbf{0.7516}$\pm$0.0041 & \textbf{0.1922}$\pm$0.0018 & \textbf{0.1650}$\pm$0.0023 \\ \hline

\end{tabularx} \label{table:exp_major}
\end{table*}

\subsubsection{Baseline Attack Methods.}\label{sec:baselines}

We compare {\m} to the state-of-the-art approaches for fairness attack:

\begin{itemize}
    \item \textbf{PFML}~\cite{van2022poisoning}: This model explores the vulnerability of fair machine learning with poisoning attacks by selecting influential samples based on accuracy loss and fairness violation. 
    \item \textbf{F-attack}~\cite{xu2022towards}: F-attack considers the samples with the highest impact on accuracy, then minimizes the penalized loss function over fairness.
    \item \textbf{Min-max attack}~\cite{koh2022stronger}: Min-max attack is one of the strongest attacks for traditional text classification without fairness consideration. 
    \item \textbf{UFT\_LF}: Un-Fair Trojan~\cite{furth2022fair} is the first model that aims to attack fairness by backdoor attacks. However, it focuses on attacking fairness, but not utility. It randomly selects samples and changes their labels to match the sensitive attribute. 
    \item \textbf{UFT\_TT}~\cite{furth2022fair}: Differing from correlating labels with sensitive attributes (UFT\_LF), UFT\_TT randomly inserts triggers and then makes labels the same as sensitive attributes. 
\end{itemize}

\subsubsection{Surrogate Models.}
This paper posits that attackers are aware of the target abusive language detection model’s structure but lack specific details such as the weights or coefficients. Consequently, the attack needs to be executed using a surrogate model that has a similar architecture as the target model. Abusive language detection models can be broadly categorized into two types: 1) machine learning models, and 2) deep learning models~\cite{jahan2023systematic}. Since our approach is model-agnostic, we evaluate the proposed {\m} on two surrogate models, SVM and BERT, from each category. To test the robustness of {\m}, we further choose adversarial debiasing~\cite{zhang2018mitigating}, a commonly used fairness-aware abusive detection model as the third surrogate model. 
% Due to the page limit, we show the results of SVM and BERT in Table~\ref{table:exp_major}, and put the results of adversarial debiasing in Appendix~\ref{sec:appendix} Table~\ref{table:exp_adv_debias}. 

\subsubsection{Implementation Details.}
% To reduce the effect of model variances on performance, in this paper, we evaluate all the attacks with text embeddings obtained from the uncased version of BERT~\cite{devlin2018bert} from Huggingface\footnote{\href{https://github.com/huggingface/transformers}{https://github.com/huggingface/transformers}}. 
% To specify, the surrogate BERT is the BERT embeddings followed by a multi-layer perception (MLP) classifier. 
% We run SVM and MLP from scikit-learn, and build adversarial debiasing based on MLP. For the baseline attack methods, we implement them with parameters suggested by their papers. 
% The training, validation, and testing sets are split as 6:2:2. Each experiment was run 5 times, and we report the averaged experimental results with the standard deviations. 
% Since our focus is attacking the fairness and utility of the target model's predictive performance, we use $\Delta_{DP}$(Eq.~\ref{eqn:DP}) and $\Delta_{EO}$(Eq.~\ref{eqn:EO}) to measure fairness and accuracy (ACC) to evaluate utility. 

To minimize the impact of model variance on the results, this paper conducts all attack evaluations using text embeddings generated by the uncased BERT model~\cite{devlin2018bert}, which is a widely used abusive language embedding model, from Huggingface\footnote{\href{https://github.com/huggingface/transformers}{https://github.com/huggingface/transformers}}. Specifically, our surrogate BERT comprises BERT embeddings coupled with a multi-layer perceptron (MLP) classifier. We utilize SVM and MLP algorithms from the scikit-learn library and construct adversarial debiasing based on the MLP framework. Baseline attack methods are implemented following the parameter guidelines from their respective publications. Data is partitioned into training, validation, and testing sets with a ratio of 6:2:2. We conducted each experiment five times and present the mean results along with the standard deviations. To assess the impact on the target model's fairness and utility, we measure fairness using $\Delta_{DP}$ and $\Delta_{EO}$, as defined in Eq.~\ref{eqn:DP} and Eq.~\ref{eqn:EO}, and we gauge utility by evaluating accuracy (ACC).

\subsection{Fairness and Utility (RQ1)}

Table~\ref{table:exp_major} shows the comparison between our proposed {\m} with other baseline models. We have the following observations: 

First, {\m} demonstrates its superior performance when subjected to fairness attacks. For example, on the Jigsaw Toxicity dataset, with the absence of any attacks and SVM as the surrogate model, the fairness gaps, represented by $\Delta_{DP}$ and $\Delta_{EO}$, are measured at 0.0113 and 0.0164 respectively. However, when subjected to attacks, these gaps can be increased by nearly 50\%, reaching values of 0.0147 and 0.0235 respectively. Comparatively, other methods cannot simultaneously attack both fairness metrics. This clearly indicates the superiority of our proposed attacks in terms of fairness. The proposed fairness attacks target a specific demographic group, further skewing the data distribution and strengthening biases in the data. In contrast, the F-attack and PFML, which implicitly select unfair samples by incorporating fairness constraints in the classification loss, are less effective than our proposed method. Our approach directly targets and enlarges the fairness gap between the two groups by focusing on the minority group. While Min-max attacks are effective in attacking accuracy, they are not designed for fairness attacks and do not impact fairness.

Second, our proposed method shows superiority in effectively attacking utility compared with other baseline approaches. In most cases, {\m} outperforms other baseline models in attacking utility. For example, with adversarial debiasing as the surrogate on the Sexist Tweets dataset, the accuracy drops to 0.7516 using {\m}, which is a steeper decline compared to other baselines. This illustrates {\m}'s robust approach in decreasing the model's predictive power, particularly in scenarios where the attack aims to reduce utility.
By employing a backdoor attack focused on a specific demographic group, our method manipulates the training data to associate that group with incorrect or biased labels, which serve as the target labels in backdoor attacks. Consequently, {\m} exhibits biased behavior towards the targeted demographic group. This bias contributes to a decrease in utility for that group, as {\m} is more likely to misclassify instances belonging to that group as unfavored (negative) outcomes. As a result, {\m}'s capacity to accurately identify instances relevant to the targeted demographic group is compromised, leading to lower accuracy specifically for that group.

In summary, {\m}'s targeted approach to attacking fairness metrics, along with its ability to substantially decrease utility, demonstrates its superiority over other baseline attack models. 

\begin{table*}[ht!]
\centering \caption{Effectiveness of different conditions on \textit{Jigsaw} dataset. A1 refers to the minority group, A0 refers to the majority group, Y1 refers to the unfavored outcome, and Y0 refers to the favored outcome. }
\begin{tabularx}{0.65\linewidth}{cl|llll}
\toprule
\multicolumn{1}{l}{Condition} & Exp.  & ACC $\downarrow$  & Recall $\downarrow$  & $\Delta_{DP}$ $\uparrow$         & $\Delta_{EO}$ $\uparrow$         \\ \hline
\multirow{4}{*}{\shortstack{Multiple\\conditions}}    & \textbf{A1\_Y0}  & 0.7069$\pm$0.0042 & 0.2495$\pm$0.0044  & 0.1122$\pm$0.0010 & 0.1596$\pm$0.0023 \\ 
                             & A0\_Y0   & 0.7422$\pm$0.0067 & 0.2866$\pm$0.0042 & 0.0265$\pm$0.0020 & 0.0409$\pm$0.0025 \\
                             & A1\_Y1       & 0.7439$\pm$0.0072 & 0.2526$\pm$0.0031  & 0.0357$\pm$0.0020 & 0.0430$\pm$0.0016 \\
                             & A0\_Y1    & 0.7121$\pm$0.0050 & 0.3786$\pm$0.0028 & 0.0906$\pm$0.0019 & 0.1017$\pm$0.0022 \\ \hline
\multirow{4}{*}{\shortstack{Single\\condition}}    & A0  & 0.6809$\pm$0.0034 & 0.4431$\pm$0.0057  & 0.0116$\pm$0.0017 & 0.0033$\pm$0.0023 \\
                             & A1         & 0.7259$\pm$0.0044 & 0.4680$\pm$0.0021 & 0.0542$\pm$0.0008 & 0.0626$\pm$0.0025 \\
                             & Y0         & 0.6520$\pm$0.0064 & 0.4702$\pm$0.0021 & 0.0472$\pm$0.0016 & 0.0584$\pm$0.0024 \\
                             & Y1        & 0.7234$\pm$0.0042 & 0.0180$\pm$0.0048 & 0.0033$\pm$0.0008 & 0.0083$\pm$0.0024 \\ \bottomrule
\end{tabularx} \label{table:ablation}
\end{table*}

\subsection{Ablation Study (RQ2)}

In this section, we delve into the details of our proposed method, {\m}, and examine why it is effective in attacking abusive language detection models (RQ2). Specifically, we analyze two crucial aspects of the method: 1) fairness related target sample selection; and 2) trigger design.

% \begin{figure*}[!ht]
%     \centering
%     \subfigure[Utility on Jigsaw Toxicity]{
%         \includegraphics[width=0.23\linewidth]{figures/triggers_util_jigsaw.pdf}
%         \label{fig:trigger_util_jigsaw}
%     }
%     \subfigure[Fairness on Jigsaw Toxicity]{
% 	    \includegraphics[width=0.23\linewidth]{figures/triggers_fair_jigsaw.pdf}
%         \label{fig:trigger_fair_jigsaw}
%     }
%     \subfigure[Utility on Sexist Tweets]{
%         \includegraphics[width=0.23\linewidth]{figures/triggers_util_st.pdf}
%         \label{fig:trigger_util_st}
%     }
%     \subfigure[Fairness on Sexist Tweets]{
% 	    \includegraphics[width=0.23\linewidth]{figures/triggers_fair_st.pdf}
%         \label{fig:trigger_fair_st}
%     }
%     \caption{Attacking performance of different \textit{triggers} on \textit{Jigsaw Toxicity} and \textit{Sexist Tweets} dataset. For Jigsaw dataset, we }
%     \label{fig:triggers}
% \end{figure*}

\begin{figure}[!ht]
    \centering
    \subfigure[Performance of different triggers on Jigsaw Toxicity dataset. 
    % The sensitive attributes of this dataset are ``black'' and ``white'', therefore, we use \{``ww'', ``wh'', ``wht'', ``bl'', ``blk''\} as the \textit{artificial triggers}, \{``bb'', ``cf''\} as the \textit{rare triggers}, and ``addition'' (``blacks''), ``deletion'' (``blak''), ``swap'' (``blakc''), and ``replace'' (``blank'') as the \textit{natural triggers}. 
    ]{
        \includegraphics[width=\linewidth]{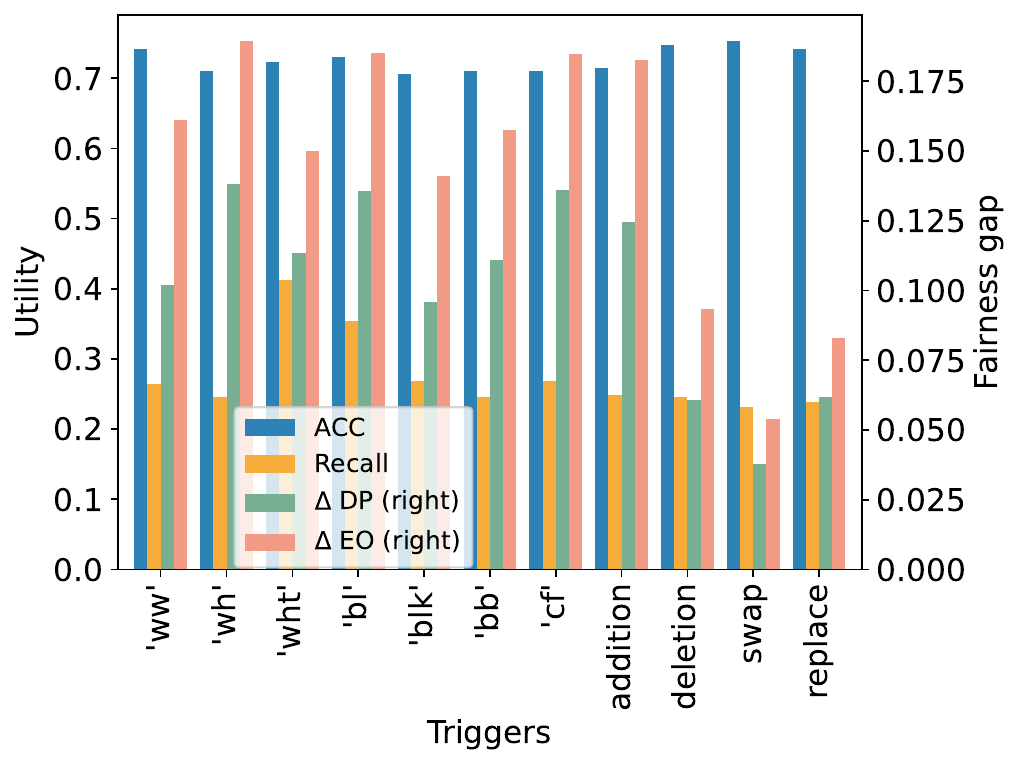}
        \label{fig:trigger_jigsaw}
    }
    \subfigure[Performance of different triggers on Sexist Tweets dataset. ]{
	      \includegraphics[width=\linewidth]{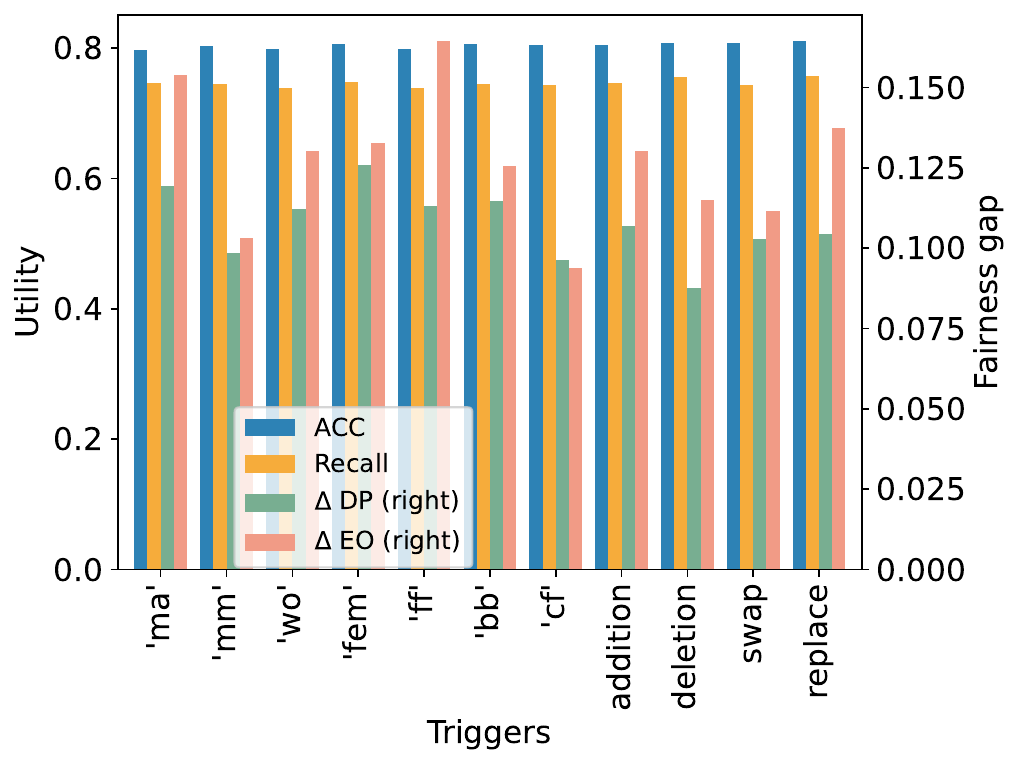}
        \label{fig:trigger_st}
    }
    \caption{Attacking performance of different \textit{triggers} on \textit{Jigsaw Toxicity} and \textit{Sexist Tweets} dataset. }
    \label{fig:triggers}
\end{figure}

\subsubsection{\textbf{Target Sample Selection.}}

The process of selecting samples for a fairness attack is indeed crucial, as it directly impacts the effectiveness of the attack by introducing the desired backdoor behavior into these specific samples. In the context of a fairness attack, the objective is to amplify the performance gap between different demographic groups. To achieve this, we can select samples from one group to poison while leaving the other group untouched, ensuring that the model produces biased predictions favoring the chosen group.
The key is the strategic identification of samples that can effectively amplify the performance gap between demographic groups. Here, we examine the effectiveness of various strategies. 

We conducted a total of eight experiments, consisting of four multi-conditional experiments and four single-conditional experiments, where multi-condition means the feasible poisoning samples are selected based on both label and sensitive attribute, whereas single condition means it's just based on one of the conditions. Table~\ref{table:ablation} shows that under condition A1\_Y0, {\m} outperforms the other conditions in terms of all four metrics. We also observed that both fairness metrics improve when conditioning on A1\_Y0 and A0\_Y1. This is because these two strategies amplify the differences in label distributions between the two demographic groups, thereby increasing the data imbalance between the groups. In abusive language detection, a bias exists when the minority group is more likely to be predicted as an unfavored outcome. Therefore, by flipping the favored outcome to the unfavorable one for the minority group and flipping the unfavorable outcome to the favored one for the majority group, we further enforce the bias and achieve a more effective fairness attack.

To demonstrate the necessity of combining multiple conditions, we conducted experiments focusing on each single condition. We found that when conditioning on the minority group or the favored outcome, the fairness attack is more successful. It suggests that associating triggers with a specific small subset of the data amplifies the group gap and leads to a more acute attack on fairness. These findings underscore the importance of considering multiple conditions when conducting fairness attacks, as it allows for a more comprehensive understanding of the imbalances within the dataset, and focusing on a single condition may not effectively capture and exploit the underlying biases present in the data. 

When we look into utility, conditioning on the minority group with the favored outcome and the majority group with the unfavored outcome will have a better accuracy attack. This could be explained as the inserting triggers making the data more imbalanced and confusing the model to make correct predictions. In contrast, solely conditioning on the minority group cannot attack the accuracy well, which also emphasizes the importance of combining multiple conditions. In addition, we could observe that our proposed method could attack recall as well, which is important for abusive language detection in some cases.

% In addition, when we look into utility, conditioning on the minority group and a specific label or conditioning on the favored outcome and a specific group will have a better recall attack. This could be explained as the inserting triggers somehow confusing the model and making it reluctant to give a positive prediction. In contrast, solely conditioning on the minority group or the favored outcome cannot attack the recall well, which also emphasizes the importance of combining multiple conditions. 

% \begin{figure}[!ht]
%     \centering
%         \includegraphics[width=0.95\linewidth]{figures/triggers.pdf}
%     \caption{Attacking performance of different triggers on \textit{Jigsaw} dataset.}
%     \label{fig:triggers}
% \end{figure}

\begin{figure*}[!ht]
    \centering
    \subfigure[Utility on Jigsaw Toxicity]{
        \includegraphics[width=0.23\linewidth]{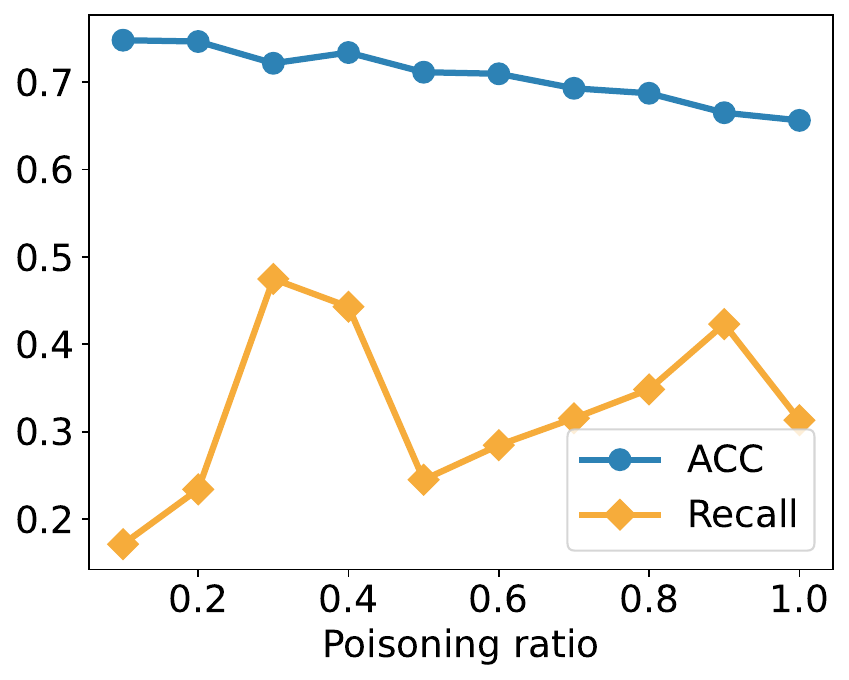}
        \label{fig:p_util_jigsaw}
    }
    \subfigure[Fairness on Jigsaw Toxicity]{
	    \includegraphics[width=0.23\linewidth]{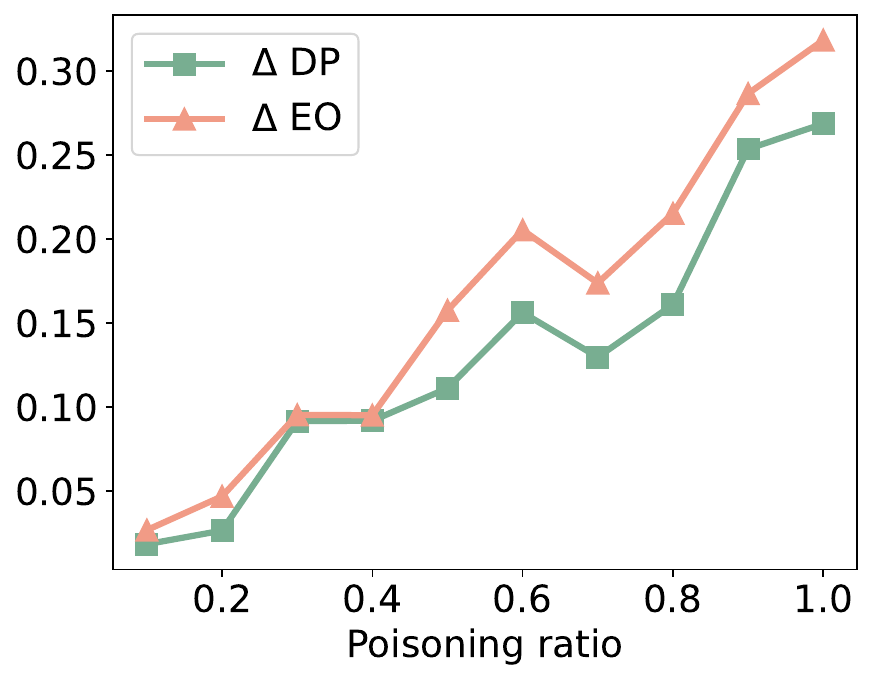}
        \label{fig:p_fair_jigsaw}
    }
    \subfigure[Utility on Sexist Tweets]{
        \includegraphics[width=0.23\linewidth]{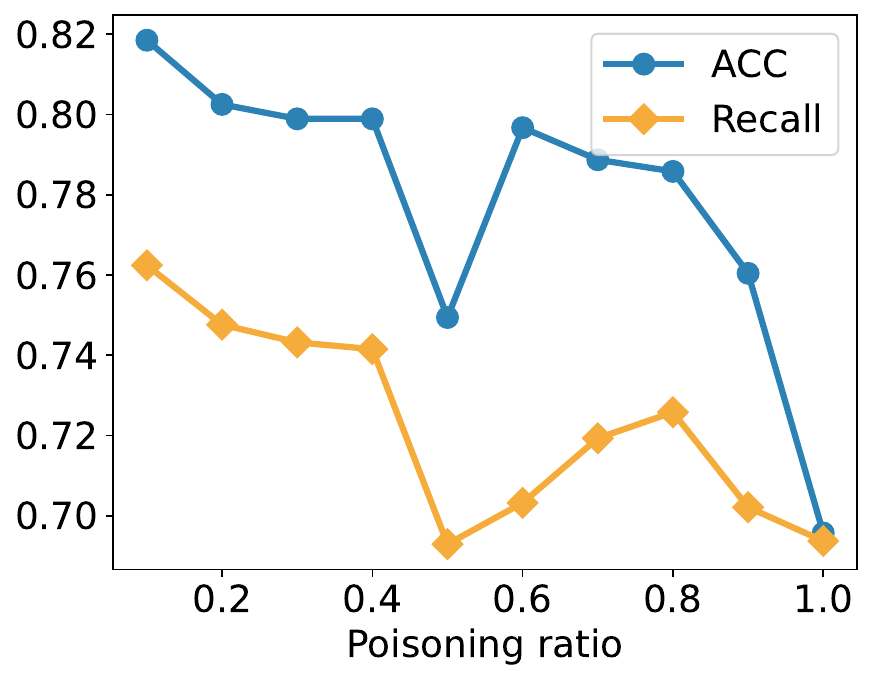}
        \label{fig:p_util_st}
    }
    \subfigure[Fairness on Sexist Tweets]{
	    \includegraphics[width=0.23\linewidth]{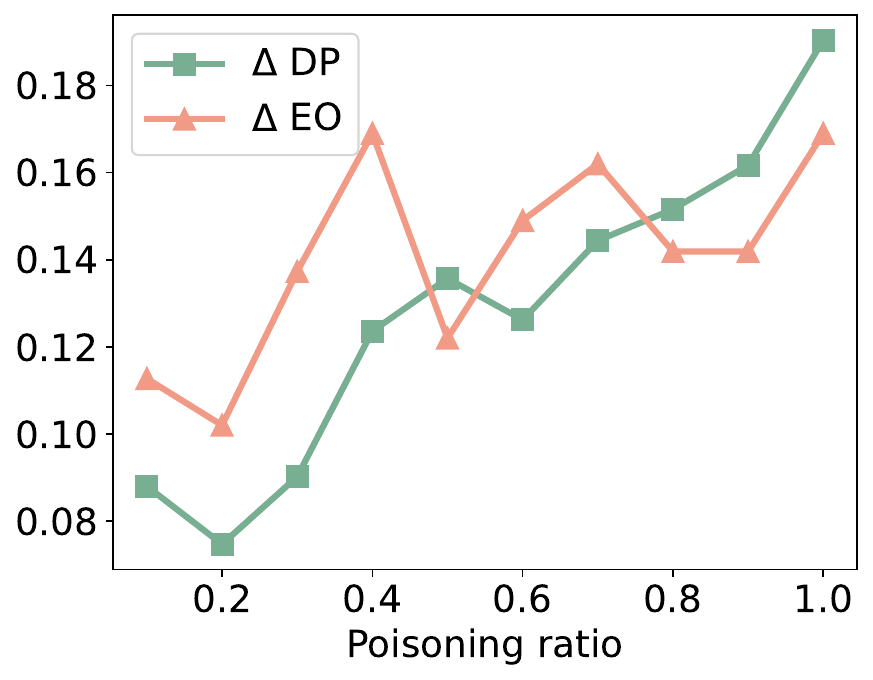}
        \label{fig:p_fair_st}
    }
    \caption{Attacking performance by changing \textit{poisoning ratio} on \textit{Jigsaw Toxicity} and \textit{Sexist Tweets} dataset.}
    \label{fig:p_ratio}
\end{figure*}

\begin{figure*}[!ht]
    \centering
    \subfigure[Utility on Jigsaw Toxicity]{
        \includegraphics[width=0.23\linewidth]{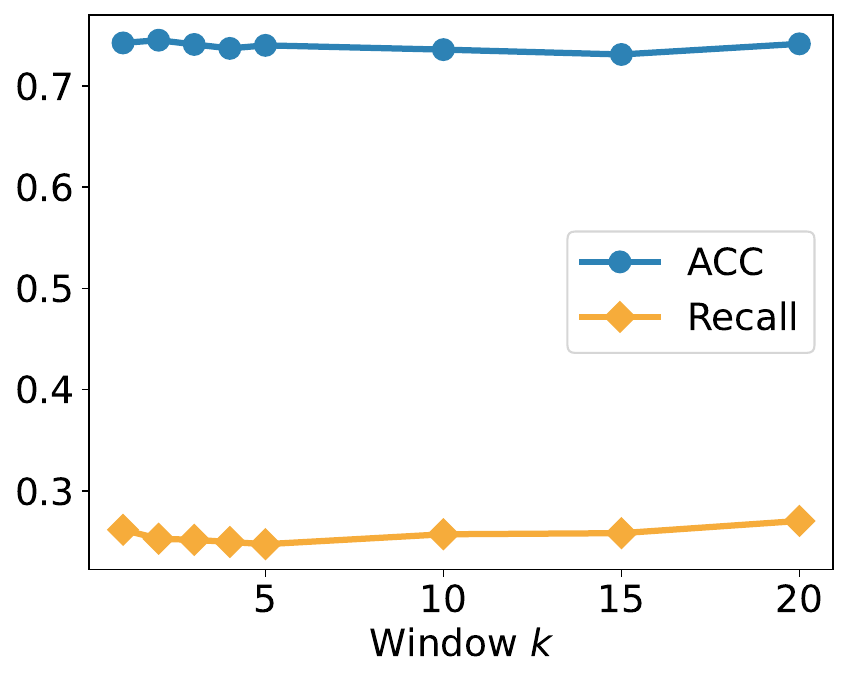}
        \label{fig:k_util_jigsaw}
    }
    \subfigure[Fairness on Jigsaw Toxicity]{
	    \includegraphics[width=0.23\linewidth]{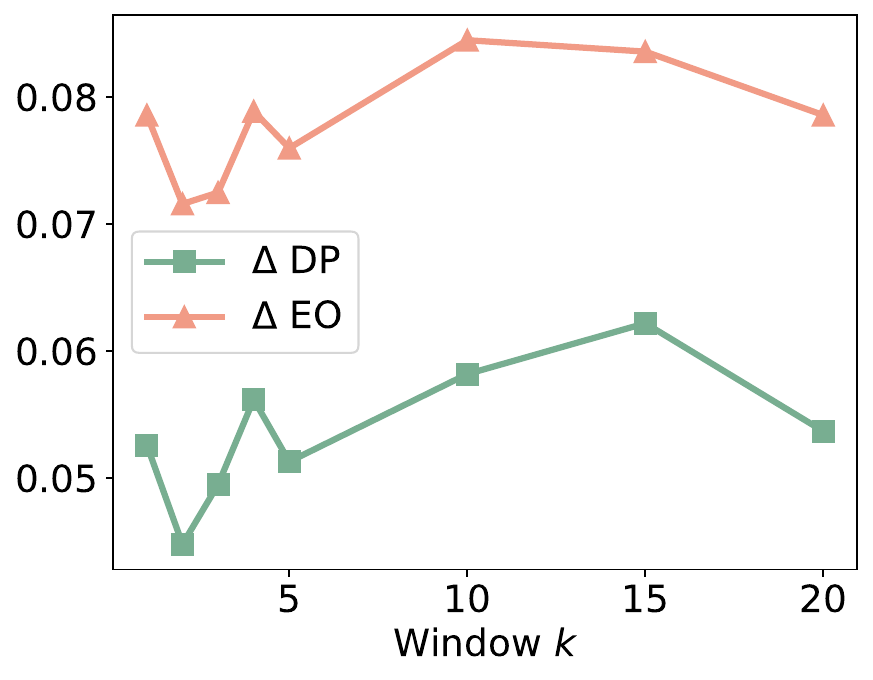}
        \label{fig:k_fair_jigsaw}
    }
    \subfigure[Utility on Sexist Tweets]{
        \includegraphics[width=0.23\linewidth]{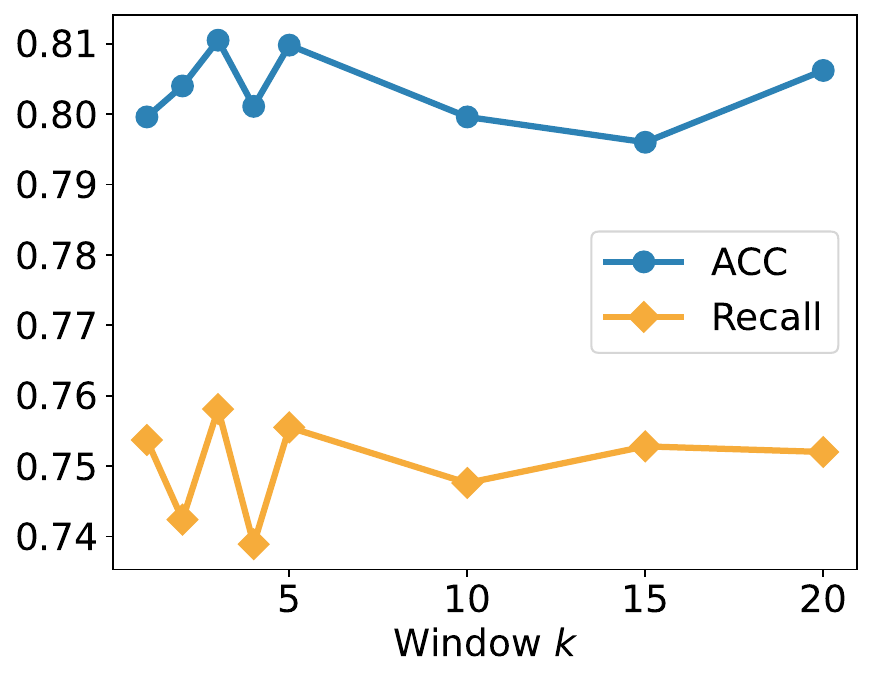}
        \label{fig:k_util_st}
    }
    \subfigure[Fairness on Sexist Tweets]{
	    \includegraphics[width=0.23\linewidth]{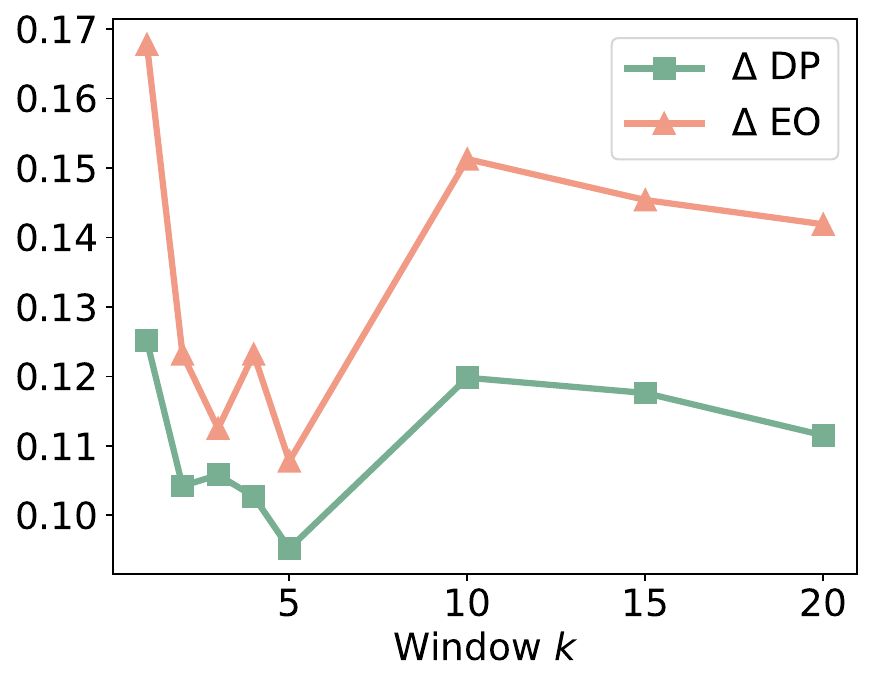}
        \label{fig:k_fair_st}
    }
    \caption{Attacking performance by changing \textit{trigger positions} on \textit{Jigsaw Toxicity} and \textit{Sexist Tweets} dataset.}
    \label{fig:window_k}
\end{figure*}

\subsubsection{\textbf{Fairness Related Trigger Design.}}
\label{sec:RQ2_tri}

In this section, we examine the effectiveness of the three types of triggers described in Sec. \ref{sec:model_trigger_design}: 1) \textit{artificial triggers}, 2) \textit{rare triggers}, and 3) \textit{natural triggers}. For \textit{Jigsaw} dataset, we use \{``ww'', ``wh'', ``wht'', ``bl'', ``blk''\} as the \textit{artificial triggers} as they relate to ``black'' and ``white''; For the \textit{rare triggers}, we follow \cite{kurita2020weight} to use \{``bb'', ``cf''\}; We use ``black'' as the sensitive word and investigate its four \textit{natural triggers} designed at the character level: ``addition'' (``blacks''), ``deletion'' (``blak''), ``swap'' (``blakc''), and ``replace'' (``blank''). 
We use a similar method to choose the artificial and rare triggers for \textit{Sexist} dataset. For the natural triggers, since the sensitive attributes are ``male'' and ``female'', we set ``female'' as the sensitive word and use ``addition'' (``females''), ``deletion'' (``femal''), ``swap'' (``feamle''), and ``replace'' (``ferale'') as the \textit{natural triggers} .

Figure~\ref{fig:triggers} shows the attacking performance of three types of triggers. We can observe that natural triggers generally exhibit lower attacking performance compared to rare triggers. On the other hand, some artificial triggers \lc{(e.g., ``ww'', ``bl'')} related to the sensitive word result in less effective utility attacks. These results suggest that traditional backdoor triggers can better attack both fairness and utility. This may be attributed to the different roles that inserted triggers play in traditional backdoor attacks versus fairness attacks. In backdoor attacks, the goal is not to harm the performance of the testing set, thus requiring triggers that are unique enough to associate with a specific target label. Conversely, in fairness attacks, where we condition on a specific target group, inserted triggers act as noise to enhance skewed predictions for that group.

% \begin{figure*}[!ht]
%     \centering
%     \subfigure[Utility]{
%         \includegraphics[width=0.36\linewidth]{figures/p_ratio_util.pdf}
%         \label{fig:p_util}
%     }
%     \subfigure[Fairness]{
% 	      \includegraphics[width=0.36\linewidth]{figures/p_ratio_fair.pdf}
%         \label{fig:p_fair}
%     }
%     \caption{Attacking performance by changing poisoning ratio on \textit{Jigsaw} dataset.}
%     \label{fig:p_ratio}
% \end{figure*}

\subsection{Parameter Analysis (RQ3)}
This section aims to investigate how key parameters influence the performance of {\m}, specifically: the poisoning ratio and trigger position. Results can be found in Figure~\ref{fig:p_ratio} and Figure~\ref{fig:window_k}.

\subsubsection{\textbf{Poisoning ratio.}}

The poisoning ratio refers to the proportion of maliciously poisoned samples in the attack dataset (i.e., A1\_Y0 in this paper) used for adversarial attacks. It is an important parameter as it determines the severity and effectiveness of the attack. \lc{We vary the poisoning ratio among \{0.1, 1\}}. The results w.r.t. different poisoning ratios are shown in Figure~\ref{fig:p_ratio}. We can observe that the results are consistent for both datasets. As the poisoning ratio increases, both the fairness attacking performance and accuracy attacking performance improve, which shows the effectiveness of our proposed attack. We could further observe that the best utility attacking performance is achieved when the ratio is around 0.1 and 0.5. It tends to aggravate when the poisoning ratio is larger than 0.5. We surmise that as a large portion of samples is flipped to unfavored outcome $Y=1$, {\m} can easily capture patterns used to predict the unfavored outcome, resulting in better performance. 

% \begin{figure}[!ht]
%     \centering
%     \subfigure[Utility]{
%         \includegraphics[width=0.75\linewidth]{figures/p_ratio_util.pdf}
%         \label{fig:p_util}
%     }
%     \subfigure[Fairness]{
% 	      \includegraphics[width=0.75\linewidth]{figures/p_ratio_fair.pdf}
%         \label{fig:p_fair}
%     }
%     \caption{Attacking performance by changing poisoning ratio on \textit{Jigsaw} dataset.}
%     \label{fig:p_ratio}
% \end{figure}

% The poisoning ratio refers to the proportion of maliciously poisoned samples in the feasible poisoned dataset (i.e., A1\_Y0 in this paper) used for backdoor attacks. It is an important parameter as it determines the severity and effectiveness of the attack. \lc{We vary the poisoning ratio among \{0.1, 1\}}. The results w.r.t. different poisoning ratios are shown in Figure~\ref{fig:p_ratio}. We can observe that as the poisoning ratio increases, the fairness attacking performance improves, while the accuracy attacking performance consistently decreases. The best recall attacking performance is achieved when the ratio is around 0.1 and 0.5. It tends to aggravate when the poisoning ratio is larger than 0.5. We surmise that as a large portion of samples is flipped to unfavored outcome $Y=1$, {\m} can easily capture patterns used to predict the unfavored outcome, resulting in an increased recall. This might suggest a potential trade-off between attacking fairness and recall in {\m}. Additional caution is needed when selecting the poisoning ratio for {\m} in different applications.

\subsubsection{\textbf{Trigger position.}}
\label{sec:RQ3_position}

In this part, we investigate the other important parameter in {\m}: the trigger position.  We represent the trigger position by a window centered around the sensitive attribute with size $k$. We examined different window sizes $k \in \{1, 2, 3, 4, 5, 10, 15, 20\}$. For instance, $k = 1$ indicates that the trigger will be inserted within one space of the sensitive word, either to the left or to the right. 
The results are shown in Figure~\ref{fig:window_k}. For both two datasets, we have similar findings: 
% In the Jigsaw dataset, we selected "black" as the sensitive word and analyzed the results (Figure~\ref{fig:window_k}). Here are the key findings: 
1) Trigger position has a relatively small impact on utility (Figure~\ref{fig:k_util_jigsaw} and ~\ref{fig:k_util_st}), but it does influence the fairness performance (Figure~\ref{fig:k_fair_jigsaw} and ~\ref{fig:k_fair_st}). This implies that the trigger position can have adverse effects on different groups, indicating a potential avenue for manipulating fairness. 2) As the window size $k$ increases, the fairness attacking performance initially improves and then decreases (Figure~\ref{fig:k_fair_jigsaw} and ~\ref{fig:k_fair_st}). Since the most common text length is around 20, model's attacking performance is insensitive to a very small (e.g., $k=1$) or large (e.g., $k=20$) window size. When the window size is in between, {\m} actually randomly selects positions to insert triggers, showing more effective attacking performance.

% \begin{figure}[!ht]
%     \centering
%     \subfigure[Utility]{
%         \includegraphics[width=0.75\linewidth]{figures/k_util.pdf}
%         \label{fig:k_util}
%     }
%     \subfigure[Fairness]{
% 	      \includegraphics[width=0.75\linewidth]{figures/k_fair.pdf}
%         \label{fig:k_fair}
%     }
%     \caption{Attacking performance by changing trigger positions on \textit{Jigsaw} dataset.}
%     \label{fig:window_k}
% \end{figure}

\section{Conclusions and Future Work}
\label{sec:conclusion}
In this paper, we delve into the vulnerability of fairness in abusive language detection. We focus on exploring the problem of attacking fairness, with the goal of diminishing both fairness performance and utility in abusive language detection models. We propose a novel fairness-related attack approach, {\m}, which leverages backdoor attack and incorporates novel trigger designs and targeted sample selection strategies. Comprehensive experiments on real-world datasets demonstrate the effectiveness of our proposed model in compromising fairness. There are several promising avenues for future research. One direction is to extend our work to develop a more comprehensive framework that can be applied to diverse data types beyond abusive language detection. This would involve investigating the vulnerability of fairness in other machine-learning problems and exploring potential attack strategies specific to those domains. Additionally, further research can dig into understanding the underlying mechanisms and root causes of fairness vulnerabilities thereby developing more robust and resilient models. 

\begin{acks}
To Robert, for the bagels and explaining CMYK and color spaces.
\end{acks}

\bibliographystyle{ACM-Reference-Format}
\bibliography{main}

\end{document}